\documentclass[10pt,twocolumn,letterpaper]{article}

\usepackage[pagenumbers]{cvpr} 

\usepackage[dvipsnames]{xcolor}

\definecolor{cvprblue}{rgb}{0.21,0.49,0.74}
\usepackage[pagebackref,breaklinks,colorlinks,citecolor=cvprblue]{hyperref}
\usepackage{booktabs}
\usepackage{colortbl}
\usepackage{amssymb}
\usepackage{booktabs}

\def\paperID{*****} 
\def\confName{CVPR}
\def\confYear{2024}

\title{Lightweight Object Detection: A Study Based on YOLOv7 Integrated with ShuffleNetv2 and Vision Transformer}

\author{Wenkai Gong\\
{\tt\small kai.901025@gmail.com}}

\begin{document}
\maketitle
\begin{abstract}
As mobile computing technology rapidly evolves, deploying efficient object detection algorithms on mobile devices emerges as a pivotal research area in computer vision. This study zeroes in on optimizing the YOLOv7 algorithm to boost its operational efficiency and speed on mobile platforms while ensuring high accuracy. Leveraging a synergy of advanced techniques such as Group Convolution, ShuffleNetV2, and Vision Transformer, this research has effectively minimized the model's parameter count and memory usage, streamlined the network architecture, and fortified the real-time object detection proficiency on resource-constrained devices. The experimental outcomes reveal that the refined YOLO model demonstrates exceptional performance, markedly enhancing processing velocity while sustaining superior detection accuracy.
\end{abstract}    
\section{Introduction}
\label{sec:intro}

\hspace{1pc}As the field of computer vision rapidly advances, object detection has become a crucial component in various applications, spanning areas such as security surveillance, autonomous driving, and smart healthcare. Despite the high computational complexity and insufficient real-time capabilities of traditional object detection methods, deep learning-based algorithms have achieved significant breakthroughs in accuracy and real-time performance. Among these, YOLO (You Only Look Once) \cite{yolo1, yolo2, yolo3, yolo4, yolo5, yolo6, yolo7, yolox}has established itself as a classic real-time object detection algorithm, striking a balance between computational speed and detection precision. However, mobile devices typically face limitations in computational power, memory capacity, and energy consumption, complicating the deployment of deep learning models. To adapt the YOLO model for these contexts, it necessitates further improvements and optimizations. This paper will delve into research on an enhanced YOLO model tailored for mobile deployment, focusing on network structure optimization, model compression and acceleration, robustness enhancement, and performance evaluation across different application scenarios.

The primary objectives of this study encompass the exploration and understanding of the YOLO algorithm and its variants in the context of object detection tasks. The focus of this work will be on grasping the fundamental principles and core mechanisms of the YOLO algorithm, along with its performance across various tasks and scenarios. This includes, but is not limited to, an in-depth investigation of YOLO's network architecture, loss functions, training strategies, and comparative analysis with other object detection algorithms. Considering the characteristics of mobile devices, this research aims to design and implement enhancements to the YOLO model. Addressing the computational capabilities and memory constraints of mobile devices, the study will strive to optimize the structure and algorithms of the YOLO model. This may involve lightweight model design, efficient algorithm implementation, and specific hardware optimizations, all intended to significantly enhance the model's performance and efficiency on mobile devices while maintaining detection accuracy. Verification and evaluation of the improved model's performance on standard datasets, as well as its operational efficiency on actual mobile devices, will also be integral. The research will further assess the performance and efficiency of the enhanced YOLO model through experimental validation on standard datasets and deployment testing in real mobile device environments. This comprehensive evaluation will help ensure that the improved model not only advances theoretically but also demonstrates feasibility and effectiveness in practical applications.

The main contributions of this paper are summarized as follows:

\begin{enumerate}
    \item In the enhanced YOLO model, the design philosophy of ShuffleNet v2 \cite{shufflenetv2} is thoroughly referenced and utilized. Particularly, the combination of channel shuffling and group convolution \cite{AlexNet} effectively balances the model's complexity and performance. This design not only improves the model's efficiency but also retains robust feature extraction capabilities, enabling real-time object detection on mobile devices. Moreover, by incorporating techniques like skip connections and depthwise separable convolutions, the model's robustness and accuracy are further enhanced.
    
    \item In the improvements made to the YOLO model, the introduction of the Vision Transformer (ViT) \cite{vit}as a core component for feature extraction not only enhances the model's ability to capture the overall image context information but also significantly improves the accuracy and efficiency in object detection. The long-range dependency capturing ability and excellent transfer learning characteristics of ViT make the model more efficient in processing complex scenes, particularly demonstrating significant real-time performance advantages in applications on mobile devices.

\end{enumerate}

\section{Related Work}
\label{sec:rel}

\subsection{ShuffleNet v2}
\hspace{1pc}ShuffleNet v2 \cite{shufflenetv2} is designed to achieve efficient computation and reduce model complexity while maintaining high performance, a challenging task since reducing complexity often risks sacrificing accuracy. However, ShuffleNet v2 successfully addresses this issue through several key innovations. Unlike its predecessor which utilized group convolutions to decrease parameters and computation, ShuffleNet v2 enhances feature interaction within each group to boost the model's representational power. Specifically, it abandons the group restriction in pointwise convolutions, allowing all channels to participate in the 1x1 convolution, simplifying the network structure, reducing memory access costs, and enhancing information flow. Additionally, by reducing channel splitting in bottleneck structures, it avoids potential information bottlenecks associated with group convolutions, achieving a more balanced computational load distribution and enhancing model efficiency. Moreover, ShuffleNet v2 optimizes the channel shuffle mechanism introduced in ShuffleNet v1 \cite{shufflenetv1}, employing ungrouped pointwise convolutions, channel splitting, and improved feature fusion strategies for more effective inter-group information exchange, thereby enriching the feature representation by reorganizing the input feature map's channel order.

\subsection{Vision Transformer (ViT)}
\hspace{1pc}The Vision Transformer (ViT)\cite{vit} is an innovative deep learning architecture designed specifically for computer vision tasks, marking a significant shift by adapting the Transformer structure, originally developed for natural language processing, to the visual domain. ViT commences by segmenting the input image into a series of patches, converting these patches into high-dimensional embedding vectors that capture localized image features. To account for the Transformer's lack of inherent sequential processing capability, positional encodings are added to these embedding vectors, enabling the self-attention \cite{Attention} mechanism within ViT to capture long-range dependencies across different image segments. The Transformer encoder processes these embeddings, focusing on various image aspects to provide robust feature representations for diverse visual tasks. The transformed vectors, particularly through a special "classification" embedding for classification tasks, are then utilized to output the final task-specific results, demonstrating ViT's adaptability and efficacy in handling complex visual information.

\subsection{You Only Look Once (YOLO)}
\hspace{1pc}Over the years, the YOLO \cite{yolo1, yolo2, yolo3, yolo4, yolo5, yolo6, yolo7, yolox} series has been one of the best single-stage real-time object detector categories. YOLO transforms the object detection task into a regression problem, predicting the positions and categories of multiple objects in a single forward pass, achieving high-speed object detection. After years of development, YOLO has developed into a series of fast models with good performance. Anchor-based YOLO methods include YOLOv4 \cite{yolo4}, YOLOv5 \cite{yolo5}, and YOLOv7 \cite{yolo7}, while anchor-free methods are YOLOX \cite{27} and YOLOv6 \cite{yolo6}. Considering the performance of these detectors, anchor-free methods perform as well as anchor-based methods, and anchor boxes are no longer the main factor limiting the development of YOLO. However, all YOLO variants generate many redundant bounding boxes, which NMS must filter out during the prediction stage, which significantly impacts the detector's accuracy and speed and conflicts with the design theory of real-time object detectors. 

\section{YOLO Model Architecture}
\label{sec:enhanced}

\subsection{Model Overview}
\hspace{1pc}This chapter focuses on introducing two key modules—the Dynamic Group Convolution Shuffle Module (DGSM) and the Dynamic Grouped Convolution Shuffle Transformer (DGST). The DGSM module is utilized for optimizing the backbone network, significantly enhancing computational efficiency while maintaining outstanding performance through the integration of group convolution and channel shuffle techniques. The DGST module, employed for optimizing the neck network, further incorporates Vision Transformer, group convolution, and channel shuffle techniques, achieving higher computational efficiency and adaptability. This module also simplifies the network structure and improves detection efficiency.

\subsection{Dynamic Group Convolution Shuffle Module (DGSM)}
\begin{figure}[htb]
    \centering
    \includegraphics[width=.8\linewidth]{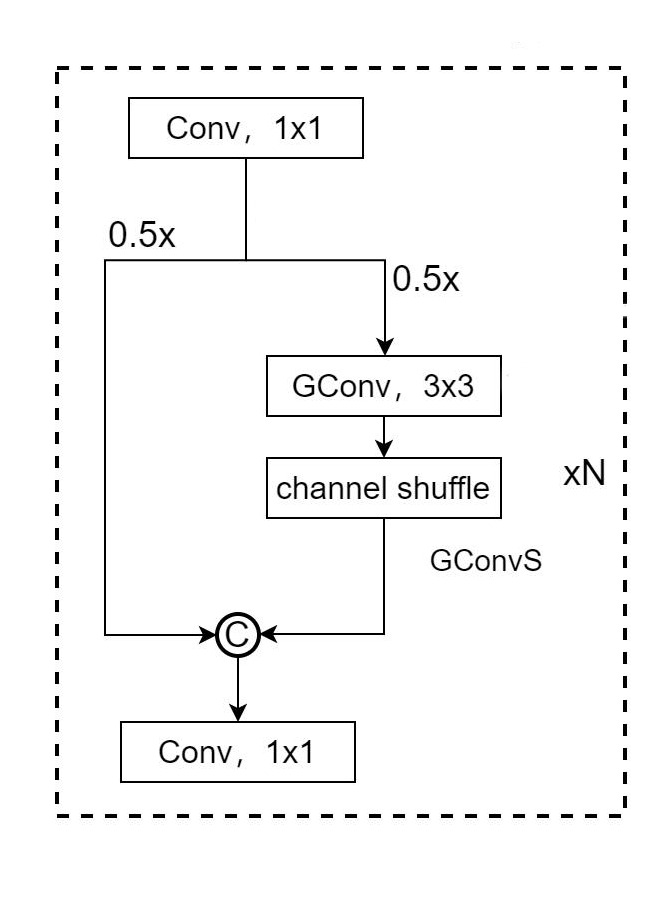}
    \caption{DGSM}
    \label{figure:DGSM}
\end{figure}

\hspace{1pc}In DGSM, \autoref{figure:DGSM} the introduction of group convolution reduces the number of model parameters and computational demands while preventing overfitting, thus maintaining the network's robustness and generalization ability. Furthermore, the channel shuffle technique from ShuffleNetV2 facilitates effective inter-group feature information exchange, crucial for preserving the network's comprehensive expressive capacity. Notably, while reducing the number of parameters, this exchange mechanism helps maintain the diversity and richness of features.

As illustrated in \autoref{table:DGSM}, the new DGSM module can precisely adjust the number of stacking blocks and the number of channels according to the requirements of different layers, replacing the original ELAN \cite{ELAN} module to form a new backbone network. This fine-tuned control and optimization approach allows the model to more effectively process features of various scales while maintaining computational efficiency, significantly enhancing the model's applicability and performance in practical applications.

\begin{table}[t]
\begin{center}
\setlength{\tabcolsep}{12pt} 
\begin{tabular}{lccccc}  
\toprule
\bf{N} &\bf{Channels} \\
\midrule
2 & 64  \\ 
\midrule
3 & 128 \\ 
\midrule
4 & 256 \\ 
\midrule
2 & 512 \\
\bottomrule
\end{tabular}%
\caption{Replacement of the Original Backbone Network's 4 ELANs with Corresponding DGSM's N and Channel Numbers.}
\label{table:DGSM}
\end{center}
\vspace{-6mm}
\end{table}

\subsection{Dynamic Group Convolution Shuffle Transformer (DGST)}

\begin{figure}[htb]
    \centering
    \includegraphics[width=.8\linewidth]{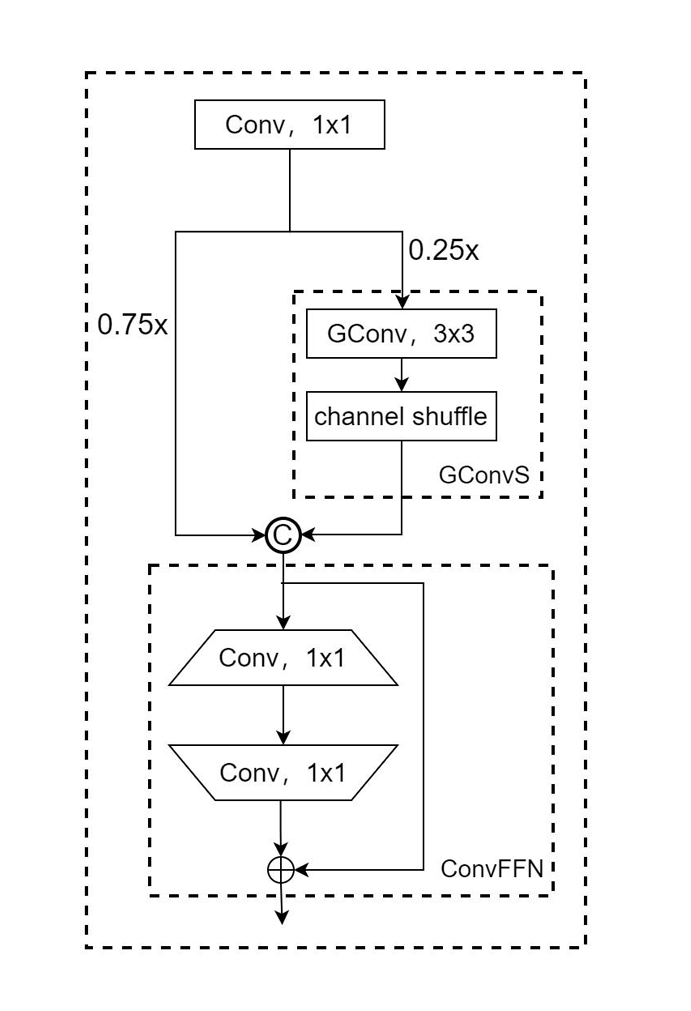}
    \caption{DGST}
    \label{figure:DGST}
\end{figure}

\hspace{1pc}The Dynamic Group Shuffle Transformer (DGST) is an innovative structure that integrates the Vision Transformer with the DGSM module, as shown in \autoref{figure:DGST}, aimed at further enhancing the model's computational efficiency and performance. The core of the DGST module is a 3:1 division strategy, where one part undergoes group convolution and channel shuffling operations, and convolution operations replace fully connected Linear to achieve the same effect, replacing this module with the original neck module. This design not only reduces computational demands but also better adapts to the characteristics of convolutional neural networks, potentially offering superior performance for the model.

\begin{figure}[htb]
    \centering
    \includegraphics[width=.8\linewidth]{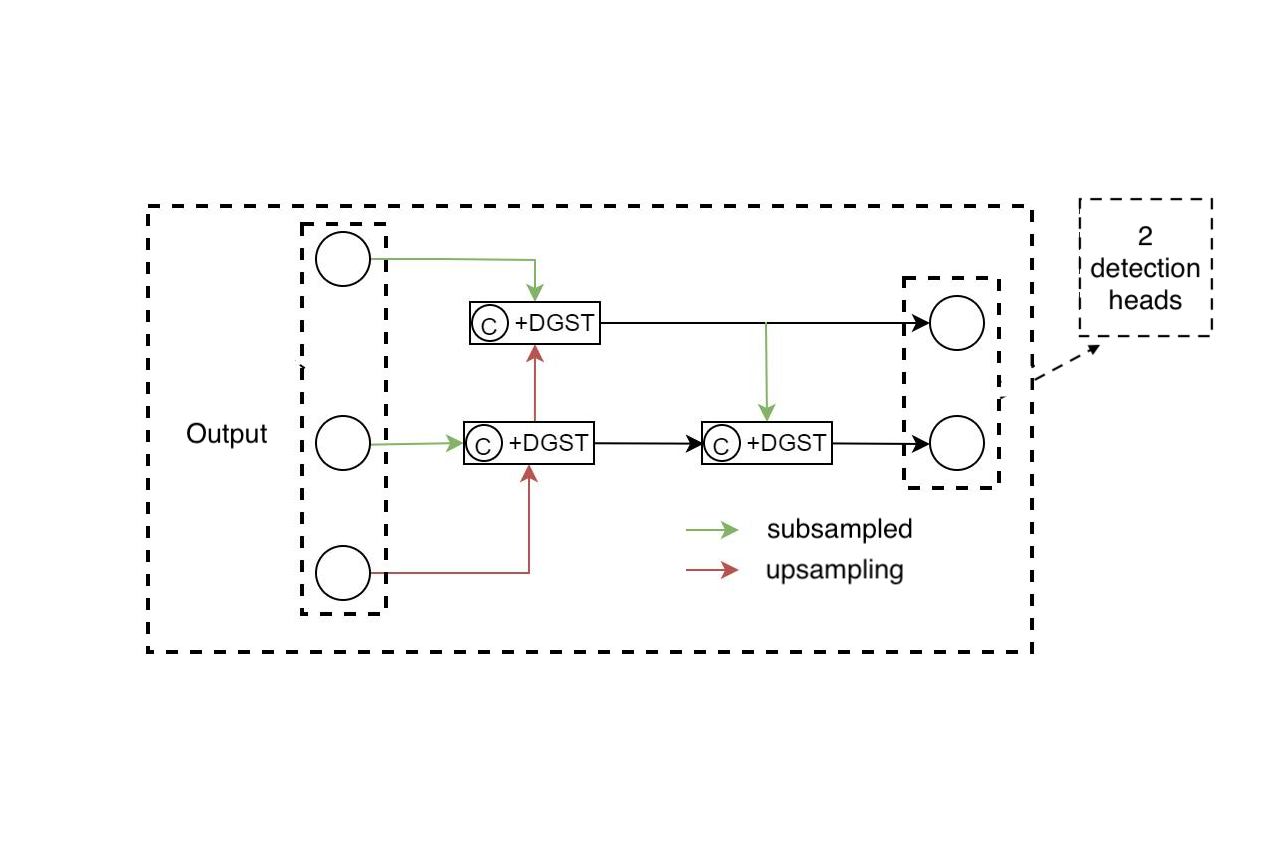}
    \caption{Reducing the original three detection heads to two}
    \label{figure:DGST_Header}
\end{figure}

To further optimize the overall network architecture, the configuration of the detection heads was adjusted, reducing the original three detection heads to two as illustrated in \autoref{figure:DGST_Header}. This modification not only alleviates the computational burden of the model but also enhances detection efficiency. The reduction in the number of detection heads implies that there is less data to process during the post-processing stage, thereby accelerating the inference speed of the entire model.
\section{Experiment}
\label{sec:exp}

\subsection{Setups}
The dataset used in this experiment comprises 1919 images of individuals, including both masked and unmasked person images. The collection of the dataset encompasses multiple sources:

\begin{itemize}
\item \textbf{Google Images}: Public images related to mask-wearing were obtained through the Google search engine.
\item \textbf{Bing Search}: Person images in various scenes and backgrounds were collected using the Bing search engine.
\item \textbf{Kaggle Dataset}: A subset of relevant images suitable for this experiment was selected from the existing datasets available on the Kaggle platform.
\end{itemize}

All images have been annotated in the YOLO format, with labels indicating whether the person is wearing a mask or not. This annotation style makes the images suitable for training the YOLO object detection model, providing a convenient data foundation for this experiment.

The strategy for dataset splitting is a crucial step that ensures effective model training and fair evaluation. The data split in this experiment is as follows:

\begin{itemize}
\item \textbf{Training Set}: Images used for model training, constituting 70\% of the dataset.
\item \textbf{Validation Set}: Images used for model tuning and hyperparameter selection, making up 15\% of the dataset.
\item \textbf{Test Set}: Images used for the final model evaluation, also representing 15\% of the dataset.
\end{itemize}

\subsection{Analysis}
As can be seen from \autoref{table:Test_loss}, the YOLOV7 Tiny model exhibits the best performance in terms of training loss, yet it also has the highest GPU consumption. The DGST+DGSM combined model offers a more balanced option when considering both GPU consumption and loss comprehensively.

\begin{table*}[htb]
\caption{Comparison of GPU Consumption and Training Loss During Training}
\label{table:Test_loss}
\centering
\begin{tabular}{lccccc}
\toprule
\bf{Method} & \bf{GPU Consumption} & \bf{Bounding Box Loss} & \bf{Object Detection Loss} & \bf{Classification Loss} & \bf{Total Loss} \\ 
\midrule
DGSM & 2.63G & 0.02671 & 0.03225 & 0.00129 & 0.06025 \\ 
\midrule
DGST & 3.52G & 0.02589 & 0.02889 & 0.001067 & 0.05585 \\ 
\midrule
DGST+DGSM & 2.33G & 0.02423 & 0.02788 & 0.0006791 & 0.05278 \\ 
\midrule
YOLOV7 Tiny & 3.79G & 0.02381 & 0.01283 & 0.0008498 & 0.0375 \\ 
\bottomrule
\end{tabular}
\end{table*}

\begin{table}[htb]
\caption{Detection Efficiency Analysis}
\label{table:test_time}
\centering
\resizebox{0.47\textwidth}{!}{%
\begin{tabular}{lccccc}
\toprule
\bf{Method} & \bf{Params(M)} & \bf{Interface(ms)} & \bf{NMS(ms)} & \bf{Total(ms)} \\
\midrule
DGSM & 4.45 & 242.1 & 1.9 & 243.9  \\ 
\midrule
DGST & 3.58 & 190.5 & 1.1 & 191.6  \\
\midrule
DGST+DGSM & 2.02 & 136.8 & 1.1 & 137.9  \\
\midrule
YOLOV7 Tiny & 6.01 & 283.4 & 1.3 & 284.7  \\ 
\bottomrule
\end{tabular}
}
\end{table}

The YOLOv7 Tiny model, as a lightweight option, demonstrated its unique performance characteristics in the experiments. During training, the GPU consumption of this model was 3.79G, the highest among the four models. Its parameter size is also the largest at 6.01M, indicating a higher model complexity. The inference time was 283.4ms and the total time was 284.7ms, both the highest among the models, which may suggest a trade-off in computational speed for implementing more complex or detailed functionalities.

The DGSM model showed certain advantages in the experiments. The GPU consumption during training was 2.63G, and the parameter size was 4.45M, indicating moderate model complexity. Its single inference time was 242.1ms, demonstrating reasonable computational efficiency. Although the total time was slightly longer at 243.9ms, this may reflect its stability in handling complex situations.

The DGST model exhibited its unique advantages in the experiment. The GPU consumption during training was 3.52G, slightly higher than that of DGSM, but its parameter size was 3.58M, slightly less than DGSM, indicating higher parameter efficiency. Its single inference time was 190.5ms and the total time was 191.6ms, both lower than DGSM, suggesting that DGST maintains good inference speed while keeping the computational burden low.

The DGSM+DGST combination model performed the best across several key indicators. The GPU consumption during training was 2.33G, relatively low, and it had the smallest parameter size at 2.02M, demonstrating excellent parameter efficiency. The inference time was 136.8ms and the total time was 137.9ms, the fastest among all models, showcasing its exceptional computational speed and efficiency.

In the further analysis of the performance of four model configurations in object detection \autoref{table:test_time}, which includes metrics such as precision, recall, and mAP, the DGST model achieved the highest F1 score (0.8524), indicating the best balance between precision and recall. The DGST+DGSM combined model followed closely with an F1 score of 0.8493, demonstrating a good balance as well.

\begin{table}[htb]
\caption{Comparison of Object Detection Performance}
\label{table:test_compare}
\centering
\resizebox{0.47\textwidth}{!}{%
\begin{tabular}{lccccc}
\toprule
\bf{Method} & \bf{Precision} & \bf{Recall} & \bf{mAP@.5} & \bf{mAP@.5:.95} & \bf{F1} \\ 
\midrule
YOLOV7 Tiny & 0.885 & 0.805 & 0.859 & 0.474 & 0.8431 \\
\midrule
DGSM & 0.895 & 0.794 & 0.849 & 0.481 & 0.8415 \\
\midrule
DGST & 0.887 & 0.806 & 0.862 & 0.475 & 0.8446 \\
\midrule
DGST+DGSM & 0.908 & 0.796 & 0.861 & 0.481 & 0.8483 \\ 
\bottomrule
\end{tabular}
}
\end{table}

The comprehensive comparison highlighted the DGST+DGSM combined model's exceptional performance in mAP@.5 and mAP@.5:.95, signifying outstanding detection capabilities across various IoU thresholds \autoref{table:test_compare} . While the DGST model showed competitive results on some metrics, the combined model exhibited more balanced and superior performance under stricter evaluation standards.
\section{Conclusion}

\hspace{1pc}When deploying object detection models on mobile devices, the primary challenges include limited computational power, memory constraints, and energy consumption issues. This study conducted thorough analyses and discussions, identifying key directions for improving lightweight models. Specifically, it successfully integrated advanced techniques such as grouped convolution, ShuffleNetV2, and Vision Transformer into the YOLOv7-tiny model, maintaining high detection efficiency while reducing model resource demands. This achievement demonstrates that with proper technology integration and innovation, the challenges of model lightweighting on mobile devices can be effectively addressed. Overall, this research has made progress in the field of lightweight real-time object detection on mobile devices, enhancing the practicality and application potential of the model in resource-constrained environments. While this work contributes theoretically and practically, the exploration in this domain remains full of challenges and opportunities, with future work focusing on optimizing model structures, improving algorithm efficiency and accuracy, and exploring the model's potential in broader application scenarios.

{
    \small
    \bibliographystyle{ieeenat_fullname}
    \bibliography{main}
}


\end{document}